%% file: root.tex
\documentclass[letterpaper, 10 pt, journal, twoside]{IEEEtran}

\IEEEoverridecommandlockouts                              %

\input{dependencies}
\input{acronyms}
\input{commands}

\title{Learning to Fly in Seconds}

\author{Jonas Eschmann$^{1,2}$, Dario Albani$^{2}$, and Giuseppe Loianno$^{1}$
\thanks{Manuscript received: November, 18, 2023; Revised March, 5, 2024; Accepted April, 3, 2024.}
\thanks{This paper was recommended for publication by Editor Aleksandra Faust upon evaluation of the Associate Editor and Reviewers' comments.
This work was supported by the Technology Innovation Institute, the NSF CAREER Award 2145277, and the DARPA YFA Grant D22AP00156-00. Giuseppe Loianno
serves as consultant for the Technology Innovation Institute. This arrangement has been reviewed and approved by the New York University in accordance with its policy on objectivity in research. }
\thanks{
$^1$The authors are with the New York University, Tandon School of Engineering, Brooklyn, NY 11201, USA. {\tt\footnotesize email: \{jonas.eschmann, loiannog\}@nyu.edu} (\textit{Corresponding author: Jonas Eschmann}).}
\thanks{$^2$The authors are with the Autonomous Robotics Research Center, Technology Innovation Institute, Abu Dhabi, UAE. {\tt\footnotesize email: \{jonas.eschmann, dario.albani\}@tii.ae}.}
\thanks{Digital Object Identifier (DOI): see top of this page.}
}

\newcommand{\insertfig}{
\includegraphics[width=0.9\textwidth]{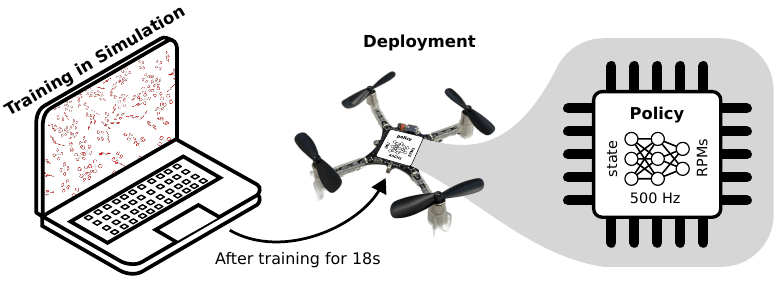}
}

\makeatletter
\apptocmd{\@maketitle}{\centering\insertfig}{}{}
\makeatother

\begin{document}

\maketitle

\input{sections/0_abstract}
\input{body.tex}
\FloatBarrier

\bibliographystyle{IEEEtran}
\bibliography{additional_references}
\end{document}

%% file: dependencies.tex
\usepackage[acronym]{glossaries}
\usepackage{graphicx}
\usepackage{placeins}
\usepackage{colortbl}
\let\labelindent\relax
\usepackage{enumitem}
\setlist[itemize]{leftmargin=0.5cm}
\usepackage{tikz}
\usetikzlibrary{tikzmark, arrows.meta, positioning}
\usetikzlibrary{shadows,calc}
\usepackage{siunitx}
\usepackage{multirow}
\usepackage{wrapfig}
\usepackage{subcaption}
\usepackage{booktabs}  
\usepackage{amsfonts}
\usepackage{xcolor}
\usepackage{siunitx}
\usepackage{mdframed}
\usepackage{adjustbox}
\usepackage{xurl}
\usepackage{pifont}
\usepackage{cite}
\makeatletter
\let\NAT@parse\undefined
\makeatother
\makeatother
\usepackage[hidelinks]{hyperref}

%% file: acronyms.tex
\newacronym{mdp}{MDP}{Markov Decision Process}
\newacronym{rl}{RL}{Reinforcement Learning}
\newacronym{dl}{DL}{Deep Learning}
\newacronym{mav}{MAV}{Micro Aerial Vehicle}
\newacronym{gpu}{GPU}{Graphics Processing Unit}
\newacronym{ml}{ML}{Machine Learning}
\newacronym{dnn}{DNN}{Deep Neural Network}
\newacronym{trpo}{TRPO}{Trust Region Policy Optimization}
\newacronym{ddpg}{DDPG}{Deep Deterministic Policy Gradient}
\newacronym{gae}{GAE}{Generalized Advantage Estimation}
\newacronym{ppo}{PPO}{Proximal Policy Optimization}
\newacronym{gps}{GPS}{Guided Policy Search}
\newacronym{cnn}{CNN}{Convolutional Neural Network}
\newacronym{rnn}{RNN}{Recurrent Neural Network}
\newacronym{relu}{ReLU}{Rectified Linear Unit}
\newacronym{tqc}{TQC}{Trucated Quantile Critics}
\newacronym{hpc}{HPC}{High Performance Computing}
\newacronym{fcu}{FCU}{Flight Controller Unit}
\newacronym{fma}{FMA}{Fused Multiply-Accumulate}
\newacronym{nvcc}{nvcc}{Nvidia CUDA Compiler}
\newacronym{stl}{STL}{Standard Template Library}
\newacronym{td3}{TD3}{Twin Delayed Deep Deterministic policy gradient}
\newacronym{vmt}{VMT}{Virtual Method Table}
\newacronym{rpm}{RPM}{Revolutions Per Minute}
\newacronym{uav}{UAV}{Unmanned Aerial Vehicle}
\newacronym{rk4}{RK4}{$4^{\text{th}}$ order Runge-Kutta}
\newacronym{gemm}{GEMM}{GEneral Matrix Multiply}
\newacronym{sfinae}{SFINAE}{Substitution Failure Is Not An Error}
\newacronym{ram}{RAM}{Random Access Memory}
\newacronym{iot}{IoT}{Internet of Things}
\newacronym{gcc}{GCC}{GNU Compiler Collection}
\newacronym{cots}{COTS}{Commercial Off-The-Shelf}
\newacronym{mems}{MEMS}{Microelectromechanical Systems}
\newacronym{vtol}{VTOL}{Vertical Take-Off and Landing}
\newacronym{pwm}{PWM}{Pulse-Width Modulation}
\newacronym{ui}{UI}{User Interface}
\newacronym{ctbr}{CTBR}{Collective Thrust and Body Rates}
\newacronym{srt}{SRT}{Single Rotor Thrusts}
\newacronym{esc}{ESC}{Electronic Speed Controller}
\newacronym{sim2real}{Sim2Real}{Simulation-to-Reality}
\newacronym{rmse}{RMSE}{Root-Mean-Square Error}
\newacronym{bpo}{BPO}{Bellman's Principle of Optimality}
\newacronym{indi}{INDI}{Incremental Non-Linear Dynamic Inversion}
\newacronym{pid}{PID}{Proportional–Integral–Derivative}
\newacronym{aac}{AAC}{Asymmetric Actor-Critic}
\newacronym{arpl}{ARPL}{Agile Robotics and Perception Lab}

%% file: commands.tex
\newcommand{\grayrow}{\rowcolor{gray!10}}
\newcolumntype{C}[1]{>{\centering\arraybackslash}p{#1}}
\usepackage{caption}

%% file: sections/0_abstract.tex
\begin{abstract}
Learning-based methods, particularly \gls*{rl}, hold great promise for streamlining deployment, enhancing performance, and achieving generalization in the control of autonomous multirotor aerial vehicles.
Deep \gls*{rl} has been able to control complex systems with impressive fidelity and agility in simulation but the simulation-to-reality transfer often brings a hard-to-bridge reality gap. Moreover, \gls*{rl} is commonly plagued by prohibitively long training times.
In this work, we propose a novel asymmetric actor-critic-based architecture coupled with a highly reliable \gls*{rl}-based training paradigm for end-to-end quadrotor control. We show how curriculum learning and a highly optimized simulator enhance sample complexity and lead to fast training times.
To precisely discuss the challenges related to low-level/end-to-end multirotor control, we also introduce a taxonomy that classifies the existing levels of control abstractions as well as non-linearities and domain parameters. Our framework enables \gls*{sim2real} transfer for direct \gls*{rpm} control after only 18 seconds of training on a consumer-grade laptop as well as its deployment on microcontrollers to control a multirotor under real-time guarantees.  Finally, our solution exhibits competitive performance in trajectory tracking, as demonstrated through various experimental comparisons with existing state-of-the-art control solutions using a real Crazyflie nano quadrotor.
We open source the code including a very fast multirotor dynamics simulator that can simulate about 5 months of flight per second on a laptop GPU. The fast training times and deployment to a cheap, off-the-shelf quadrotor lower the barriers to entry and help democratize the research and development of these systems. 
\end{abstract}

%% file: body.tex
\section*{Supplementary Material}
\noindent \textbf{Video}: \href{https://youtu.be/NRD43ZA1D-4}{\url{https://youtu.be/NRD43ZA1D-4}}

\noindent \textbf{Code}: \href{https://github.com/arplaboratory/learning-to-fly}{\url{https://github.com/arplaboratory/learning-to-fly}}

\noindent \textbf{Parameters}: \{\href{https://github.com/arplaboratory/learning-to-fly}{\textbf{Code}}\} {$\rightarrow$} \texttt{./media/parameters.pdf}
\input{sections/1_introduction}
\input{sections/2_taxonomy}
\input{sections/3_related_work}
\input{sections/4_approach}

\input{sections/5_results}

\input{sections/6_conclusion}

%% file: sections/1_introduction.tex
\section{Introduction}
\IEEEPARstart{W}{ith} the availability of cheap \gls*{cots} gyroscopes and accelerometers that are implemented as \gls*{mems}, the large-scale production of cheap \glspl*{uav}, particularly quadrotors, became viable. Bearing \gls*{vtol} as well as hovering capabilities a myriad of use cases, such as search and rescue, infrastructure inspection, or package delivery emerged. Leveraging classical, cascaded control hierarchies, multirotors are able to perform a variety of tasks. However,  these control approaches require domain expertise and engineering to be adapted to new platforms and use cases. 
At the same time, the recent developments in machine learning and particularly the success of deep learning for supervised tasks like image classification \cite{krizhevsky_imagenet_2012, he_deep_2015}, raise the question if these learning-based capabilities could be transferred to quadrotor control. In contrast to supervised learning, (multirotor) control is a decision-making problem that can be phrased as a \gls*{mdp} where labels usually do not directly exist. To solve \glspl*{mdp}, \gls*{rl} has been employed to train policies for complex end-to-end continuous control problems in simulation \cite{mnih_human-level_2015, lillicrap_continuous_2016, schulman_high-dimensional_2015, schulman_proximal_2017}.
However, while the results attained in simulation are impressive, transferring end-to-end control policies to real-world systems has proven challenging. This is mainly due to model inaccuracies, partial observation of the state, observation and action noise, and other disturbances. Additionally, \gls*{rl} is well known to be sensitive to the choice of hyperparameters and reward function design. The requirement for hyperparameter tuning and reward function design, in combination with long training times for end-to-end control as shown in Table \ref{table:related_work_training_times}, can prohibit fast iteration and create a barrier to entry. 
To better highlight the challenges and clarify the scope of true end-to-end control, we describe the different levels of quadrotor dynamics and control by developing a taxonomy to categorize related work.
We argue that the abstractions in classic control stacks introduce information loss (e.g., actuator constraints in differential flatness-based control) and constrain the expressivity of the control policy. Due to \gls*{bpo}, the optimal policy for a decision/control problem can be represented as a function mapping from states/observations to actions. Hence, in theory, the optimal policy can be expressed as a neural network which in the limit is a general function approximator \cite{hornik_multilayer_1989}. Compared to general function approximators, classical control stacks (based on multi-level abstractions) have limited expressivity and can not necessarily represent the optimal policy, especially when accounting for evermore (non-linear) factors in the simulation. Optimization-based controllers can circumvent this limitation but are usually too computationally expensive to run end-to-end control under real-time constraints on microcontrollers. Hence, we believe it can be desirable to design an end-to-end controller using \gls*{rl}, directly mapping the quadrotor state to \gls*{rpm} outputs.  

We observe that the end-to-end control of quadrotors using state-of-the-art deep \gls*{rl} techniques (especially using off-policy \gls*{rl}) is not well explored and documented. It is particularly unclear which level of performance end-to-end policies can achieve compared to classic controllers when deployed directly on a real quadrotor under real-time constraints. With this work, we aim to push the boundaries of deep \gls*{rl}-based end-to-end quadrotor control and present the following contributions
\begin{itemize}[noitemsep]
\item \textbf{\gls*{rl}-based end-to-end controller design}: We propose a novel asymmetric actor-critic-based architecture coupled with a highly reliable training paradigm for true end-to-end quadrotor control (level \textbf{5.1} outputs, cf. Fig. \ref{fig:taxonomy} and Table \ref{table:related_work_training_times}). The proposed training paradigm takes advantage of the ground truth available in the simulator while explicitly accounting for the partial observability of the real system using an action history.

\item \textbf{Best sample complexity}: We devise a curriculum that gradually increases the penalties in the reward function leading to better sample complexity and more reliable policies. We show the benefit of the components in our proposed training paradigm by conducting an extensive ablation study containing \num{300} real-world trajectories across different configurations, seeds and tasks. In contrast to existing works, our training setup uses off-policy \gls*{rl} and we demonstrate the training of an end-to-end quadrotor control policy using the fewest number of environment interactions reported.
\item \textbf{Fastest training time}: By implementing a highly optimized simulator, we demonstrate the fastest training of an end-to-end quadrotor control policy that can be transferred to a real system. 
\item \textbf{Sim2Real}: We conduct extensive experiments including more than $300$ flights across configurations, seeds and tasks to test the \gls*{sim2real} transfer of end-to-end policies for direct \gls*{rpm} control. We show that training a position controller using our setup generalizes to other tasks like (agile) trajectory tracking. 
\item \textbf{Open Source}: We open-source our setup to facilitate research, enabling everyone with a consumer-grade laptop to train and deploy state-of-the-art quadrotor control policies in a matter of seconds, greatly reducing the barriers to entry in this research area and therefore contributing to democratize the use of these approaches.
\end{itemize}

%% file: sections/2_taxonomy.tex
\section{Multirotor Dynamics and Control Taxonomy}
\label{sec:taxonomy}
\input{figures/taxonomy.tex}
In the following, we introduce a taxonomy classifying the different abstraction levels in multirotor dynamics and control. We believe this taxonomy to be beneficial to the discussion as it allows for the precise categorization of different controllers, particularly those that we analyze in the related work (see Section~\ref{Section3:RelatedWork}). Furthermore, it explicitly exposes at which levels non-linearities and domain parameters exert an influence on the motion of multirotors.
We list the different control levels based on the order of the system when expressed in terms of a flat state \cite{mellinger_minimum_2011}. The taxonomy shown in Fig. \ref{fig:taxonomy} is expressed from the perspective of a position controller and each subsequent level denotes a lower level of control inputs. The sub-bullets denote non-linear transformations that allow expressing the dynamical system using different inputs. These transformations are not only challenging because of the non-linearity they bear but also due to the additional system/domain parameters they might introduce (marked in square brackets). 
It is worth noting that every layer incorporates an additional level of indirection, manifesting in the form of integrators. 
The lower the input level to the system is chosen, the more detached the effect on the higher level (e.g., the position) is from the cause (e.g., \gls*{rpm} setpoints). For controllers on the lowest levels, the cause-effect relationship traverses through up to five orders of integration and multiple non-linear transformations that are dependent on system parameters.
We would like to highlight that \textbf{3.1 Angular rate \& thrust} inputs are commonly referred to as ``low-level'' commands but from the taxonomy, we can see that there is only one domain parameter (mass) and otherwise just the rotational kinematics of a rigid body. Alternatively, \textbf{3.1 Angular rate \& thrust} inputs are also commonly referred to as \gls*{ctbr} inputs. Compared to the \gls*{ctbr} level, the reality gap and complexity of the system greatly increases towards the lowest levels like \textbf{5.1 Motor commands/\gls*{rpm} setpoints} which is the level of control used in this work. The high order of integration between inputs and the desired output (position) poses a great challenge for \gls*{rl} algorithms because the high-frequency exploratory actions (from $\epsilon$-greedy-like exploration schemes) are already suppressed in the early layers and hence lead to high-sample complexity and unreliable training behavior. In our proposed architecture, we overcome this challenge by using a combination of off-policy \gls*{rl}, curriculum learning, and by scheduling the exploration noise. 
From Fig. \ref{fig:taxonomy}, we can observe that the complexity of the control problem and the size of the reality gap, follows a superlinear scaling where most of the non-linear dynamics and system parameters can be found in the lower levels. 

%% file: figures/taxonomy.tex
\begin{figure}[htbp]
\begin{enumerate}[noitemsep,leftmargin=*,labelindent=0.6cm]
    \item[\tikzmarknode{a0}{0.}] \textbf{Position}
    \item[\tikzmarknode{a1}{1.}] \textbf{Velocity}
    \item[\tikzmarknode{a2}{2.}] \textbf{Acceleration}
\begin{enumerate}
    \item[\tikzmarknode{a21}{2.1}] \textbf{Attitude/orientation \& thrust}: Non-linear orientation $\rightarrow$ acceleration transfer function [mass]
\end{enumerate}
    \item[\tikzmarknode{a3}{3.}] \textbf{Jerk}
\begin{enumerate}
    \item[\tikzmarknode{a31}{3.1}] {\textbf{Angular rate \& thrust (CTBR)}: Non-linear rotational kinematics}
\end{enumerate}
    \item[\tikzmarknode{a4}{4.}] \textbf{Snap}
\begin{enumerate}
    \item[\tikzmarknode{a41}{4.1}] \textbf{Body torque  \& thrust}: Rotational dynamics [inertia]
    \item[\tikzmarknode{a42}{4.2}] \textbf{Individual rotor thrusts (SRT)}: [vehicle geometry]
    \item[\tikzmarknode{a43}{4.3}] \textbf{\Glspl{rpm}}: Non-linear torque/thrust curves [torque/thrust model parameters]
\end{enumerate}
    \item[\tikzmarknode{a5}{5.}] \textbf{Crackle}
\begin{enumerate}
    \item[\tikzmarknode{a51}{5.1}] \textbf{Motor commands/\Gls{rpm} setpoints}: [first-order low-pass time constant (motor delay)]
    \item[\tikzmarknode{a52}{5.2}] \textbf{Motor effort}: [battery level]
\end{enumerate}
\end{enumerate}
\begin{tikzpicture} [overlay, remember picture, >=Stealth]
    \draw [->, line width=1pt] let \p1=(a0.north), \p2=(a52.south) in ([xshift=-1.2em]\x1, \y1) -- node[midway, left, rotate=90, anchor=center, yshift=0.8em] {Difficulty \& Uncertainty} ([xshift=-1.2em]\x1, \y2);
    \draw[red] (current bounding box.south west) rectangle (current bounding box.north east);
\end{tikzpicture}
\vspace{-10pt}
\caption{Taxonomy of multirotor dynamics and control.
}
\label{fig:taxonomy}
\vspace{-5pt}
\end{figure}
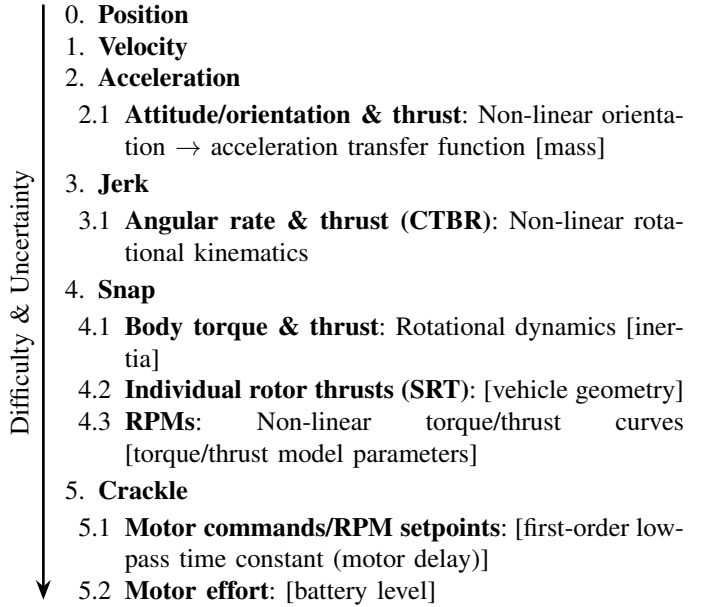

%% file: sections/3_related_work.tex
\section{Related Works}
\label{Section3:RelatedWork}
\textbf{Simulators}.
In the past, a large number of general robotics and specifically quadrotor simulators have been proposed. Many of these focus on visual fidelity and photo-realistic reproduction of the environment (e.g., to enable realistic RGB camera perception) \cite{hutchison_hector_2014, hutter_airsim_2018, song_flightmare_nodate, panerati2021learning}. Others, as the one presented in this work, focus on the accurate implementation of quadrotor dynamics \cite{Furrer2016, rotortm2022, song_flightmare_nodate, panerati2021learning}. Among the referenced simulators, Flightmare \cite{song_flightmare_nodate} is the most related because it emphasizes the simulation speed of the quadrotor dynamics for the sake of \gls*{rl}. 
Our simulator also focuses on fast dynamics but since we are concerned with low-level control we just provide a basic \gls*{ui} instead of the photo-realism that is offered by other simulators.

\textbf{Reinforcement learning for quadrotor control.}
As described in Section~\ref{sec:taxonomy}, it is usually easier to learn controllers on the higher levels, especially for Sim2Real transfer. 
Hence, a lot of work that focuses on downstream tasks has been using \gls*{ctbr} and higher-level control inputs to train \gls*{rl} based agents. The use of velocity commands has been particularly common~\cite{sadeghi_cad2rl_2017, polvara_autonomous_2018, sampedro_image-based_2018, belkhale_model-based_2021, rubi_deep_2021, kaufmann_benchmark_2022}. Fewer works also use orientation level commands \cite{becker-ehmck_learning_2020, lin_flying_2019} or angular rate commands (\gls*{ctbr}) \cite{kooi_inclined_2021,kaufmann_deep_2020, kaufmann_benchmark_2022, kaufmann2023champion}. The latter is usually used when increased agility is required (e.g., for acrobatics, racing, or flying through a narrow gap). We would like to highlight that, while impressive, recent work on learned, champion-level drone racing \cite{kaufmann2023champion} is using higher-level control outputs (level \textbf{3.1 \gls*{ctbr}}) and heavily relies on classic lower-level controllers to bridge the \gls*{sim2real} gap. In their work, the policy is only exposed to rotational kinematics and uncertainty about the mass, which is easy to identify (cf. Table \ref{table:related_work_training_times}, Fig. \ref{fig:taxonomy}). Our end-to-end policy (level \textbf{5.1 Motor commands}) is faced with the full stack of non-linearities and uncertainties about the dynamics parameters.

To use the quadrotor dynamic's full potential, the focus has been shifting to training agents outputting individual rotor thrusts (also referred to as \gls*{srt}) \cite{hwangbo_control_2017,molchanov_sim--multi-real_2019, pi_low-level_2020,song_autonomous_2021,kaufmann_benchmark_2022,gronauer2022usingsimulation}. The authors in~\cite{zhang_learning_2023} even go lower-level, training a controller outputting \gls*{rpm}. These represent the works that are closely related to our proposed solution. In the following, we discuss their similarities and differences. One of the earliest demonstrations of the successful application of deep \gls*{rl} for quadrotor control has been presented in \cite{hwangbo_control_2017}. The authors train a position controller that outputs individual rotor thrusts but in contrast to our approach, they are using a complex training procedure (exploration scheme) that requires a resettable simulator and drives up the sample complexity to more than \num{100} million environment steps. Additionally, they do not take into account rotor delays and their code is not available. In \cite{molchanov_sim--multi-real_2019}, the authors apply domain randomization to transfer a policy that outputs \gls*{srt} to different quadrotors but they require knowledge about the particular thrust-to-weight ratio and thrust limits. In comparison, our approach demonstrates \gls*{sim2real} transfer without domain randomization and without modifications to the state estimation in the firmware as well as much faster training times (comparison in Table \ref{table:related_work_training_times}). 
\begin{table}
\def\arraystretch{1.3}
\centering
\begin{tabular}{p{1.4cm} C{0.7cm} C{0.5cm} C{1.4cm}}
\hline
Publication & Level & Time & Samples \\
\hline
\grayrow
2019 \cite{molchanov_sim--multi-real_2019} ${}^\dagger$         & 4.2/5.1 & \SI{30}{\hour}  & $84 \times 10^6$  \\
2020 \cite{pi_low-level_2020}          & 4.3     & N/A  & $10 \times 10^6$ \\ 
\grayrow
2021 \cite{song_autonomous_2021}                   & 4.3     & $\sim~\SI{2}{\hour}$ & N/A  \\
2021 \cite{kooi_inclined_2021}                     & 2.1     & $27 \; \mathrm{m}$ & $1 \times 10^6$  \\
\grayrow
2022 \cite{gronauer2022usingsimulation} ${}^\dagger$ & 4.2/5.1     & $\SI{1.2}{\hour}$ & $16 \times 10^6$ \\
2023 \cite{kaufmann2023champion}                   & 3.1     & $50 \; \mathrm{m}$ & $100 \times 10^6$  \\
\grayrow
\textbf{Ours}                                      & \textbf{5.1}     & $\mathbf{18 \; \mathrm{s}}$ &  $\mathbf{0.3 \times 10^6}$  \\
\hline
\end{tabular}
\caption{Training times. {4.2/5.1}: Policy outputs 4.2, simulation 5.1. ${}^\dagger$: Used as a baseline in Table \ref{table:controller_comparison}.}
\label{table:related_work_training_times}
\end{table}
In \cite{kaufmann_benchmark_2022} the authors perform a benchmark of training controllers outputting velocity, \gls*{ctbr} and \gls*{srt} commands but in contrast to our work they do not manage to transfer the low-level controller (\gls*{srt}) to the real world.
Moreover, in all of the aforementioned works as well as in \cite{pi_low-level_2020,song_autonomous_2021,kaufmann_benchmark_2022,gronauer2022usingsimulation}, \gls*{srt} control outputs are used which simplify the learning problem by not exposing the agent to the non-linear \gls*{rpm}~$\leftrightarrow$~thrust relationship. 
Lastly in \cite{zhang_learning_2023}, \gls*{rpm} outputs are used but rather than cutting out lower-level controls to simplify the learning problem as commonly done and described in Section \ref{sec:taxonomy}, they discard the higher-level control and take \gls*{ctbr} as input from a high-level controller.
In addition, in contrast to our method, \cite{zhang_learning_2023} requires a state history and domain randomization as well as training for \SI{2}{\hour} and \num{100}M steps. Furthermore, \cite{zhang_learning_2023} only works under near-hover conditions and fails for agile trajectories (unlike ours, cf. Table~\ref{table:controller_comparison}, Fig. ~\ref{fig:trajectory_tracking_lissajous_3.5} and the video). Lastly, the complexity of the method and the lack of an open-source implementation inhibit replication.
In contrast to all the most related work which rely on on-policy, policy gradient \gls*{rl} algorithms (particularly \gls*{ppo} \cite{schulman_proximal_2017}), we use \gls*{td3}\cite{fujimoto2018addressing}, an off-policy \gls*{rl} algorithm which offers better sample complexity and achieves very fast wall-clock training times.

\label{sec:approach}
\begin{figure*}[htbp]
    \centering
    \includegraphics[width=\linewidth]{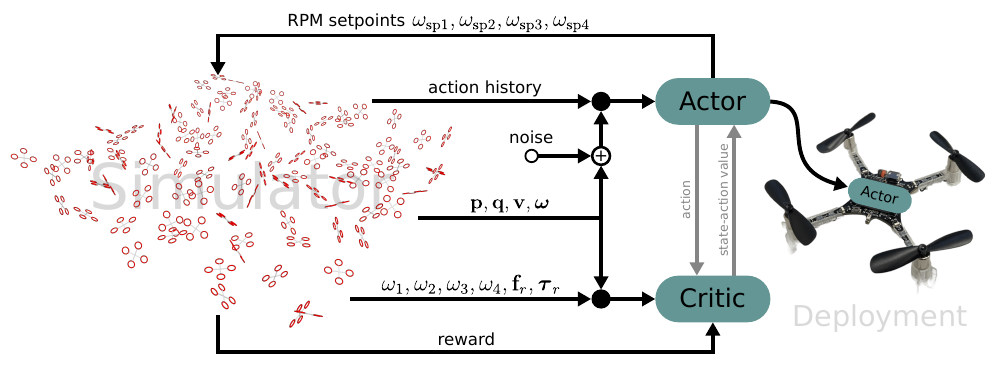}
    \caption{Overview of the training and inference setup with a view of the simulator UI (left).}
    \label{fig:overview}
    \vspace{-15pt}
\end{figure*}

%% file: sections/4_approach.tex
\section{Methodology}
To be able to take full advantage of the capabilities of the robot,  we phrase our control problem as an \gls*{mdp} where the policy directly maps states to motor commands in the form of \glspl*{rpm}. 
We select quaternions to represent orientation as they are a compact and global representation.
However, for the observations fed into the actor and critic, we convert them to rotation matrices to remove the ambiguity stemming from the quaternion's double coverage of the space of rotations. 
Since we model the motors as a first-order low-pass filter, \glspl*{rpm} are also part of the state. This makes the state $17$ dimensional with the following structure $\mathbf{s} = \{\mathbf{p}, \mathbf{q}, \mathbf{v}, \boldsymbol{\omega}, \boldsymbol{\omega}_{m}\}$, consisting of position, orientation, linear and angular velocity and motor speeds respectively. 
We implement the standard dynamics of a quadrotor subject to motor delays (please refer to e.g. \cite{song_flightmare_nodate, kaufmann2023champion, kaufmann_benchmark_2022} or our parameter reference link).
On real platforms, motor speeds are usually not observable and \gls*{rpm} setpoints are fed to the \gls*{esc} in a feed-forward, one-way fashion through \gls*{pwm}.
Hence, we implement an asymmetric actor-critic scheme \cite{pinto_asymmetric_2017}, where the critic, being only required during training, has access to privileged information from the simulator as shown in Fig.~\ref{fig:overview}. In particular, we let the critic access the \glspl*{rpm} and a random force $\mathbf{f}_r$ and torque $\boldsymbol{\tau}_r$ disturbance that are sampled at the beginning of each episode to increase the robustness of the trained policies. 
Hence, the privileged observations of the critic are represented by a \num{28} dimensional vector $\mathbf{o_c} = \{
\mathbf{p},\mathbf{R},\mathbf{v},\boldsymbol{\omega}, \boldsymbol{\omega}_m,\mathbf{f}_r,\boldsymbol{\tau}_r\}$ consisting of position, orientation, linear velocity, angular velocity, rotor speeds, force disturbance, and torque disturbance respectively. 

We also notice that the delay in the step-response is significant and caused by the motors' low-pass behavior. This behavior also strongly impacts the dynamics of the \SI{27}{\gram} nano quadrotor (Crazyflie) used in this work. Common values for the RC equivalent time constant ($1-e^{-1} \approx 63\%$ step response) are between \SI{0.05}{\second} and \SI{0.25}{\second}. We empirically find \SI{0.15}{\second} to be suitable for \gls*{sim2real} transfer which is also supported by the manufacturers measurements\footnote{Crazyflie motor step response: \url{https://web.archive.org/web/20220309092320/https://www.bitcraze.io/wp-content/uploads/2015/02/M1-step-response.png}}. These delays are significantly larger than the usual control interval of low-level controllers which usually run at around hundreds of Hz. Therefore, these delays lead to actions only impacting the state after \num{5} to \num{25} control steps. To mitigate this large level of partial observability, we add a history of control actions to the actor's observation. The action history is a proprioceptive measurement that can be trivially implemented in software on any real-world platform without requiring additional hardware (in contrast to direct \gls*{rpm} feedback measurements). Therefore, the actor's observations are $18+N_H\cdot4$ dimensional, where $N_H$ is the length of the action history, and are defined as follows: $\mathbf{o_a}=\{
\mathbf{p},\mathbf{R},\mathbf{v},\boldsymbol{\omega},\mathbf{H}\}$ with $\mathbf{H}$ being the action history. 

While the critic observes the ground truth state, the actor's observations are additionally perturbed using observation noise to account for imperfections in the sensors (the scale of the noise components is described with all other parameters in the supplementary material). The actions consist of the \gls*{rpm} setpoints as $\mathbf{a}=\{\omega_{\text{sp}1},\omega_{\text{sp}2}, \omega_{\text{sp}3}, \omega_{\text{sp}4}\}$ which places our policy/controller on the lowest order (level 5.1) of the taxonomy described in Section \ref{sec:taxonomy}. 

For the initial state distribution, we sample from a diverse set of positions, orientations, linear and angular velocities as well as rotor speeds.
We use a negative squared cost with an additive constant incentivizing survival to mitigate the ``learning to terminate'' problem \cite{eschmann2021reward}
\begin{align*}
r(s, a, s') = &  - C_{rp}\lVert \mathbf{p}\rVert_2^2 - C_{rq}\left(1-{q_w}^2\right) - C_{rv}\lVert \mathbf{v}\rVert_2^2 \\
& - C_{r\omega}\lVert \mathbf{\boldsymbol{\omega}}\rVert_2^2 - C_{ra}\lVert\mathbf{a} - C_{rab}\rVert_2^2 + C_{rs}.
\end{align*}
The values of the constants $C_*$ are supplied in the supplementary material. Additionally, we find that a simple curriculum that is transitioning from an initial set of constants $C_{\text{init},*}$ to a more restrictive $C_{\text{target},*}$ (punishing position errors and particularly control actions more harshly) benefits the sample complexity and \gls*{sim2real} transfer (as described in Section \ref{Section5:Setup}). Every \num{100000} steps the weights are adjusted by multiplying them by the $C_{p*}$ factors described in the supplementary material until they reach the $C_{\text{target},*}$ values, where they remain constant. We also decay the exploration noise using the same exponential scheme as in the curriculum of the reward function. 

We would also like to highlight that we train a position controller, not merely a stabilizing controller that only works around a particular state. The goal of our position controller is to return to the origin point with zero linear velocity from any initial conditions (within reasonable bounds described in the parameters in the supplementary material). Hence, we train a policy that can go to any position or velocity setpoint by shifting the current position and velocity (so there is no need to apply goal-conditioned \gls*{rl}). For stable behavior, the position and velocity errors induced by the shifted setpoints should not exceed the errors seen during training which can easily be accommodated for by clipping.

To facilitate fast training times we implement a highly optimized simulator for multirotor dynamics. A sample view of the interface is shown in Fig.~\ref{fig:overview} (left). By leveraging C++ template metaprogramming the simulation code can be highly optimized by the compiler and can be tightly integrated into the RLtools~\cite{eschmann2023rltools} deep \gls*{rl} framework.

%% file: sections/5_results.tex
\section{Experimental Setup}\label{Section5:Setup}

\textbf{Simulation}.
We run our multirotor dynamics simulator on a Nvidia T2000 laptop GPU and attain \num{1284} million steps/s. To reach this level of performance, \num{64} blocks of \num{128} threads are each executing the forward dynamics in parallel. This amounts to \num{8192} environments in total which are run for \num{1000000} steps each. The required execution time of the GPU kernel is \SI{6380}{\milli\second} amounting to \num{1284} million steps/s. At a simulation frequency of \SI{100}{\hertz} this is about \num{5} months of simulated flight per second. Compared to Flightmare \cite{song_flightmare_nodate} which is the state of the art in terms of dynamics simulation speed with a reported frequency of \num{200000} steps/s on a laptop, our simulation is about $6420\times$ faster. 

\begin{figure}[h]
\centering
\includegraphics[width=\linewidth]{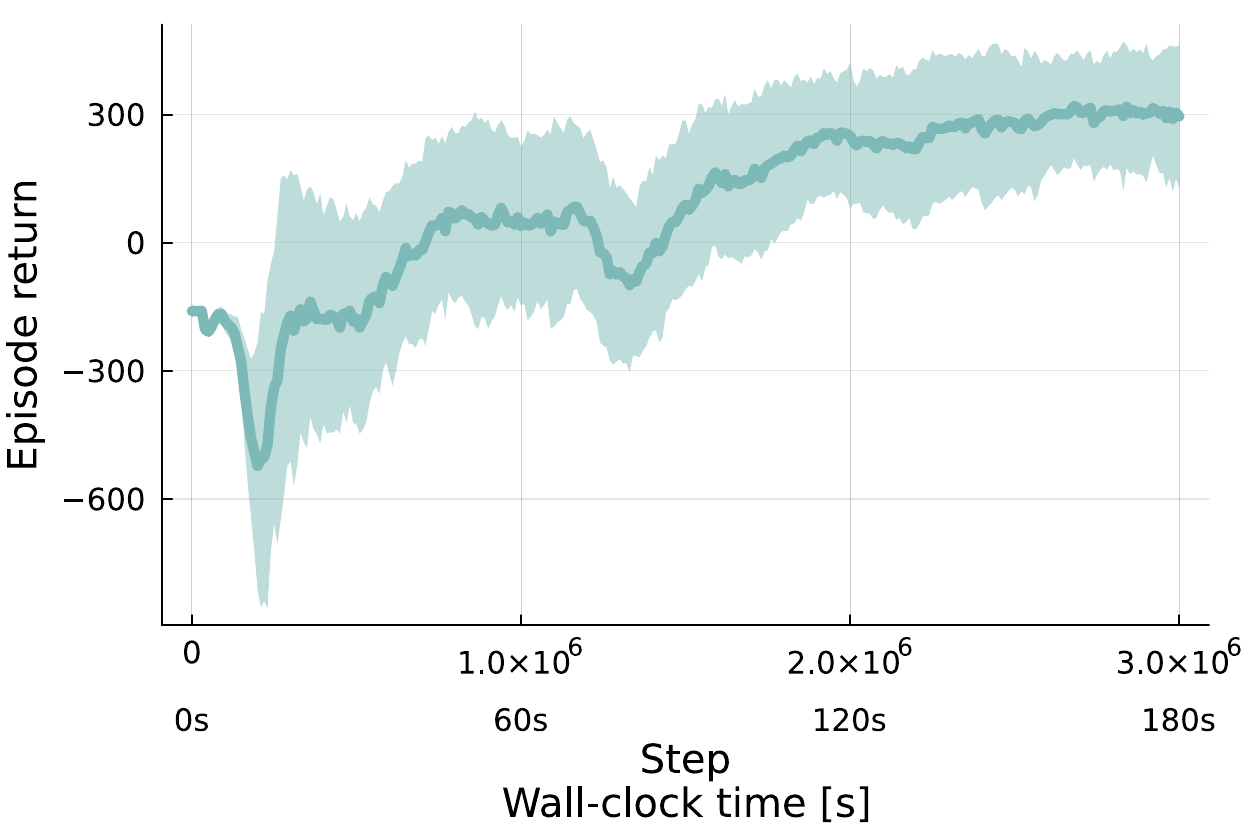}
\caption{Episode return ($\mu$ and $\sigma$ over $50$ runs with different initial seeds).}
\label{fig:learning_curves_returns_aggregate}
\end{figure}

\begin{figure}[h]
\includegraphics[width=\linewidth]{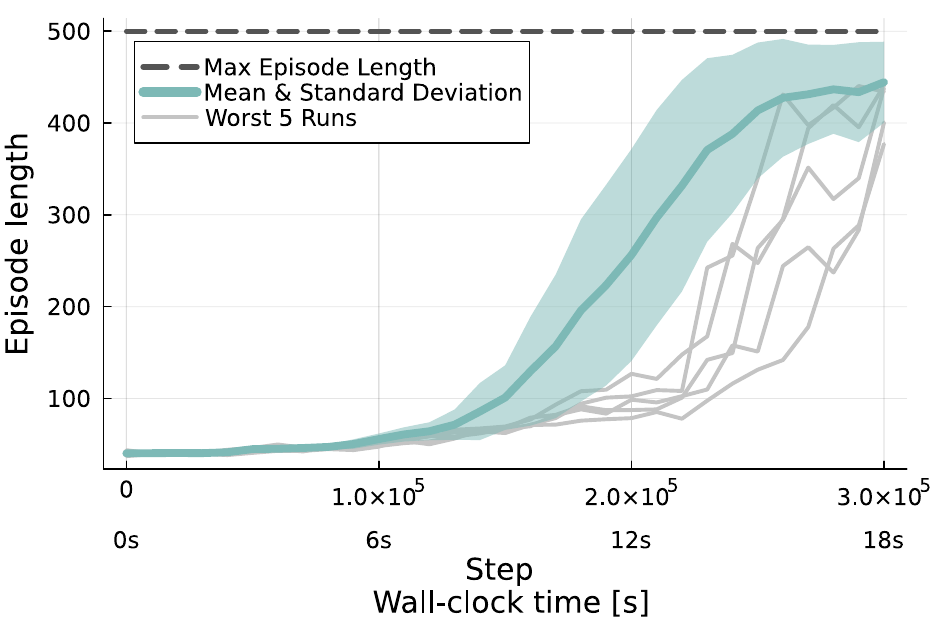}
\centering
\caption{Episode lengths ($\mu$ and $\sigma$ over $50$ runs with different initial seeds). Note, compared to Fig. \ref{fig:learning_curves_returns_aggregate} the horizontal axis is zoomed in to highlight the initial phase.}
\label{fig:learning_curves_episode_lengths_aggregate_zoom}
\end{figure}

\textbf{Training}.
Leveraging the training setup described in Section \ref{sec:approach} and using the RLtools \gls*{rl} framework \cite{eschmann2023rltools}, we train a low-level quadrotor control policy. Fig. \ref{fig:learning_curves_returns_aggregate} shows the learning curve in terms of the returns (sum of rewards per episode). Purely looking at the reward/returns, it is hard to judge the level of flying capabilities of a particular policy because it is highly dependent on the reward function formulation. We believe the episode length is a more understandable measure than the return because if the agent crashes or flies away from a tight box around the origin, the episode is terminated. Hence, from Fig. \ref{fig:learning_curves_episode_lengths_aggregate_zoom} we can see that after about \num{300000} steps (total number of interactions with the environment) or \SI{18}{\second} of training on a 2020 MacBook Pro, the policy has learned to fly relatively reliably. 
In comparison to related work (listed in Table \ref{table:related_work_training_times}) our approach is substantially faster and requires an order of magnitude less samples when compared to learned policies on a similar level of control outputs.

We investigate the $50$ differently seeded training runs by selecting the $5$ seeds with the worst cumulative number of steps over the whole training run (area under the curve) shown in Fig. \ref{fig:learning_curves_episode_lengths_aggregate_zoom}. We notice that even the worst $5$ runs learn to fly rapidly. This shows the remarkable reliability of our training approach which is not commonly expected from \gls*{rl} training pipelines (e.g., in \cite{molchanov_sim--multi-real_2019} the authors state that cherry-picking across many seeds is required to find a policy that can fly). This remarkable level of reliability is also confirmed by real-world experiments that are described in the following.
\input{tables/ablation_study}

\section{Results}
\label{Section6:Results}
\textbf{Ablation study.} To show the impact of the different components of the training setup (as shown in Fig. \ref{fig:overview}) on the performance during training and real-world deployment, we conduct a large-scale ablation study and present the results in Table \ref{table:appendix_ablation_study}. To account for the inherent stochasticity of the observation noise and training process, we take a moving average over the previous 10 evaluations for the episode return and length metrics. The evaluation takes place every 1000 steps, hence the moving window covers 10000 steps. We train each configuration for up to \num{3000000} steps using \num{50} different initial seeds each. Figs. \ref{fig:learning_curves_returns_aggregate} and \ref{fig:learning_curves_episode_lengths_aggregate_zoom} are generated from the \num{50} runs of the baseline configuration (containing all components). Furthermore, we execute the resulting policy after \num{300000} and \num{3000000} steps of training using the first \num{10} different seeds of each configuration on a real Crazyflie quadrotor.  
We manually terminate the position control episodes after \SI{20}{\second} of flying because in our experience, at this level, policies can fly without crashing until the battery runs out. In the case of the trajectory tracking task, we complete $4$ subsequent cycles or report a failed attempt if the quadrotor crashes beforehand. 
We report the mean and median as well as the minimum of the position error across the \num{10} runs/seeds. The minimum positional error is a particularly interesting metric because it corresponds to cherry-picking which is common in related works (e.g., \cite{molchanov_sim--multi-real_2019}). 
Since the curriculum entails changes in the reward function and we apply reward recalculation of all rewards in the replay buffer after each modification of the reward function, we also ablate the setup without reward recalculation. 
Note that the configurations differ in the added/removed complexity. The components observation noise, reward recalculation, exploration noise decay, and disturbances are minor changes, while the asymmetric actor-critic structure, action history as well as rotor delay, and the curriculum are larger modifications.

From the results in Table \ref{table:appendix_ablation_study}, we can see that overall (and particularly when executed on the real system) the baseline is the most reliable with no crashes in any of the tasks/seed combinations. In general, we can observe a trend that removing the smaller modifications still yields reliable policies (early during the training as well as after convergence). In contrast, when removing the more complex components, we can see a more pronounced drop in reliability as well as tracking performance. Here, we can observe that the curriculum indeed strongly impacts the training speed. Without the curriculum, the policies are not able to fly early on while after convergence they reach a comparable performance to the baseline. We can observe a similar behavior when removing the \gls*{aac} and hence also ablate removing both, the asymmetric actor-critic and the curriculum and find that the training takes considerably longer and even after \num{3000000} steps most of the policies are crashing. 
As deduced from first principles, we can also confirm that training without simulating the rotor delay leads to no usable policy. 
When simulating the rotor delays but not including the action history we can still see a considerable drop in reliability and positional accuracy, particularly after \num{3000000} steps where the policy seems to be overfitting the system dynamics. This validates the need for an action history to account for the partial observability in our training setup.
When taking into account the average episode lengths and returns achieved by the different configurations during training in simulation we can only observe a loose correlation between simulation and real-world performance. As described earlier, the episode lengths appear to be a better gauge for the real-world performance as well.

We conclude that overall the baseline configuration with all components and the configurations with minor ablations yield the best performance when taking into account the speed of training (sample complexity) and position error. We observe that the baseline configuration yields the most reliable policies at all stages and anecdotally is also most robust with respect to e.g. turbulent wind (cf. the supplementary video). For trajectory tracking, after \num{3000000} steps in particular, we find that the configuration without exploration noise gives the lowest tracking error among the tested seeds and hence chose it for the trajectory tracking experiments in the next section.

\input{figures/lissajous_tracking}

\begin{figure}[htb]
\centering
\includegraphics[width=1.0\linewidth]{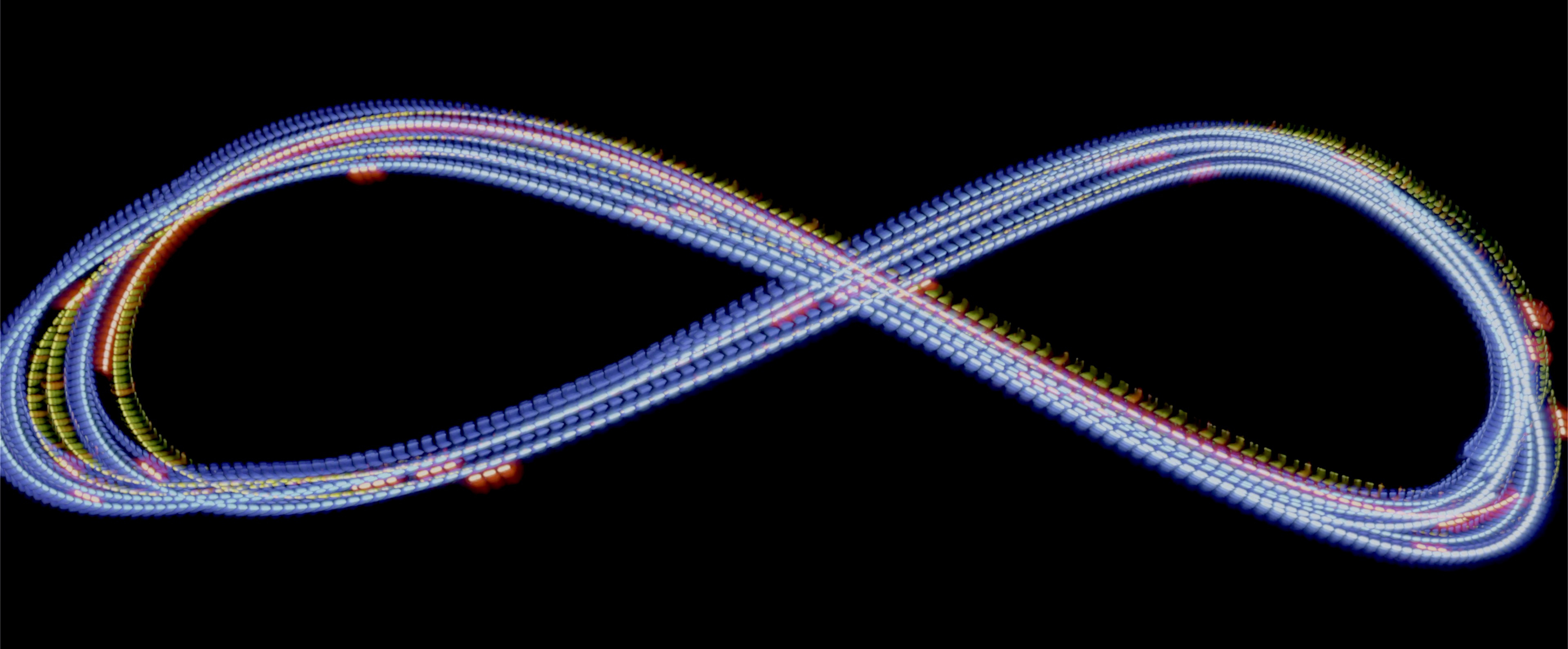}
\caption{Long exposure photo of the real-world tracking of a Lissajous trajectory with a $5.5~\mathrm{s}$ cycle time.}
\label{fig:trajectory_tracking_lissajous_5s_long_exposure}
\end{figure}

\input{tables/trajectory_tracking_results}

\textbf{Trajectory tracking.} In addition, even though we train a position controller, we find that the resulting policies can track trajectories like the Lissajous in Fig. \ref{fig:trajectory_tracking_lissajous} when deployed on the real quadrotor. This is obtained by feeding into the policy an offset. We  test the tracking performance using a figure-eight Lissajous trajectory $\mathbf{p}(t) = \begin{bmatrix}
\cos(2\pi t/T) & \sin(4\pi t/T)/2 & \text{const}\end{bmatrix}^\top$ with varying cycle times $T$ and show the tracking performance in Fig. \ref{fig:trajectory_tracking_lissajous}.
We conduct the experiments in a flying space of $10\times6\times4~\si{m^3}$ at the \gls*{arpl} at New York University (equipped with a Vicon motion capturing system).
We can see that our trained policy is able to track even agile trajectories like in Fig. \ref{fig:trajectory_tracking_lissajous_3.5} where it reaches up to $\SI{3}{\metre\per\second}$ and accelerations of up to $0.9$~g. In Fig. \ref{fig:trajectory_tracking_lissajous_5s_long_exposure}, we show a long-exposure photo of the trajectory in Fig. \ref{fig:trajectory_tracking_lissajous_5.5} being tracked by one of our policies.

We execute the same trajectory using different types of classical controllers: \gls*{pid}, geometric \cite{mellinger_minimum_2011}, nonlinear \cite{BrescianiniNonlinearController2013} and \gls*{indi} \cite{smeur_indi_2016}.
We report the results in Table \ref{table:controller_comparison}. Additionally, we also compare the proposed solution to additional RL baselines. We replicate the related works \cite{molchanov_sim--multi-real_2019} and \cite{gronauer2022usingsimulation}. In accordance with the original work (cf. Table \ref{table:related_work_training_times}), we train them for \num{84000000} and \num{16000000} steps respectively. In the case of \cite{molchanov_sim--multi-real_2019}, we find that none out of $10$ seeds leads to a policy that can fly the real drone. We extracted a trained checkpoint from the firmware provided by the authors. We found this checkpoint to be able to fly the trajectories but exhibit major oscillations and tracking error for the normal and fast trajectories. Based on the authors' instructions we believe this checkpoint was cherry-picked across even more seeds. In comparison, we found the method by \cite{gronauer2022usingsimulation} to train more reliably and to result in low position error for the slow trajectory. On the other hand, the policies trained using \cite{gronauer2022usingsimulation} exhibited poor yaw control (spinning around z while tracking the trajectory) and crashed for the normal-speed trajectory for 7/10 seeds and for 10/10 seeds in the fast case. We also test training \cite{molchanov_sim--multi-real_2019} and \cite{gronauer2022usingsimulation} for only \num{300000} and \num{3000000} steps for $10$ seeds each (same as our method) but none of the resulting policies were able to fly. Across the classic controllers, we find that for slow trajectories, the geometric controller~\cite{mellinger_minimum_2011} achieves lower tracking error than our method. While for normal-speed trajectories our method is on par with the best classic controllers, it outperforms other approaches in the fast case.

%% file: tables/ablation_study.tex
\begingroup
\setlength{\tabcolsep}{4.5pt}
\begin{table*}
\vspace{5pt}
\def\arraystretch{1.3}
\centering
\begin{tabular}{c | c c c c | c c c c c c c c c c c c}
 & \multicolumn{4}{c|}{Training (Simulation)} & \multicolumn{12}{c}{Inference (Real World)} \\
\hline
Task & \multicolumn{4}{c|}{Position Control} & \multicolumn{4}{c|}{Position Control} & \multicolumn{8}{c}{Trajectory Tracking} \\
\hline
Checkpoint [steps] & \multicolumn{2}{c|}{\num{300000}} & \multicolumn{2}{c|}{\num{3000000}} & \multicolumn{8}{c|}{\num{300000}}  & \multicolumn{4}{c}{\num{3000000}}\\
\hline
Ablation & $N$ & $R$ & $N$ & $R$ & 
\# & $\overline{e}$ & $e_{\text{med}}$ & $e_{\text{min}}$ &
\# & $\overline{e}$ & $e_{\text{med}}$ & $e_{\text{min}}$ &
\# & $\overline{e}$ & $e_{\text{med}}$ & $e_{\text{min}}$ 
\\
\hline
\grayrow
  \textbf{\small{All Components}} & 	444 & -171 & 	366 &  296 & \textbf{\small{10/10}} & 0.26 & 0.24 & 0.1 & \textbf{\small{10/10}} & 0.34 & 0.38 & \textbf{\small{0.18}} & \textbf{\small{10/10}} & 0.27 & 0.26 & 0.21 \\
  Observation Noise & 	446 & -195 & 	364 &  294 & \textbf{\small{10/10}} & 0.26 & 0.27 & 0.1 & 8/10 & 0.33 & 0.31 & 0.23 & \textbf{\small{10/10}} & \textbf{\small{0.21}} & 0.21 & 0.16 \\
\grayrow
  Reward Recalculation & 	442 & -311 & 	397 & \textbf{\small{325}} & 9/10 & \textbf{\small{0.24}} & \textbf{\small{0.15}} & \textbf{\small{0.07}} & 9/10 & 0.34 & \textbf{\small{0.3}} & 0.25 & 7/10 & 0.22 & 0.22 & 0.2 \\
  Exploration Noise Decay & 	444 & -171 & 	351 &  285 & 9/10 & 0.25 & 0.22 & 0.08 & 8/10 & \textbf{\small{0.31}} & \textbf{\small{0.3}} & 0.19 & \textbf{\small{10/10}} & \textbf{\small{0.21}} & \textbf{\small{0.18}} & \textbf{\small{0.15}} \\
\grayrow
  Disturbances & \textbf{\small{446}} & \textbf{\small{-146}} & 	379 &  316 & 9/10 & 0.28 & 0.27 & 0.08 & 8/10 & 0.33 & 0.32 & 0.21 & \textbf{\small{10/10}} & \textbf{\small{0.21}} & 0.21 & 0.17 \\
  Asymmetric Actor-Critic & 	290 & -313 & 	431 &  268 & 6/10 & 0.28 & 0.29 & 0.08 & 4/10 & 0.4 & 0.34 & 0.32 & 9/10 & 0.25 & 0.26 & 0.19 \\
\grayrow
  Action History & 	173 & -355 & 	300 &   38 & 5/10 & 1.18 & 1.41 & 0.25 & 5/10 & 0.63 & 0.6 & 0.25 & 0/10 & $\infty$ & $\infty$ & $\infty$ \\
  Curriculum & 	444 & -340 & \textbf{\small{450}} & -285 & 1/10 & 0.18 & 0.18 & 0.18 & 1/10 & 0.23 & 0.23 & 0.23 & 9/10 & 0.2 & 0.2 & 0.15 \\
\grayrow
  Rotor Delay &  46 & -207 &  44 & -195 & 0/10 & $\infty$ & $\infty$ & $\infty$ & 0/10 & $\infty$ & $\infty$ & $\infty$ & 0/10 & $\infty$ & $\infty$ & $\infty$ \\
  AAC \& Curriculum & 	296 & -341 & 	428 & -540 & 0/10 & $\infty$ & $\infty$ & $\infty$ & 0/10 & $\infty$ & $\infty$ & $\infty$ & 2/10 & 0.26 & 0.26 & 0.23 \\
\hline
\end{tabular}
\caption{Ablation study: The \textit{Ablation} column describes the component that is removed,  except for the first row which contains all components. Returns $R$ and number of steps per episode ($N$) are mean over \num{50} training runs with different seeds each. The mean($\overline{e}$)/median($e_{\text{med}}$)/minimum($e_{\text{min}}$) of the position error (in the xy-plane) are statistics over \num{10} flights of the real quadrotor with policies from different seeds each (no cherry-picking, seeds are the first \num{10} of the \num{50} trained in simulation). The \textit{\#} column shows the number of successful flights (without crashing). For the trajectory tracking task we use the figure-eight trajectory shown in Fig. \ref{fig:trajectory_tracking_lissajous_5.5} with an interval of $T=\SI{5.5}{\second}$. For each metric, the best value is marked in bold. The best values of the real-world tests are awarded provided that at least $5/10$ of the runs/seeds of a particular configuration are successful for each of the three tests.}
\label{table:appendix_ablation_study}
\vspace{-20pt}
\end{table*}
\endgroup

%% file: figures/lissajous_tracking.tex
\begin{figure*}[htbp]
    \centering
    \captionsetup[subfigure]{skip=-3pt}
    \begin{subfigure}{0.33\textwidth}
        \centering
        \includegraphics[width=\linewidth]{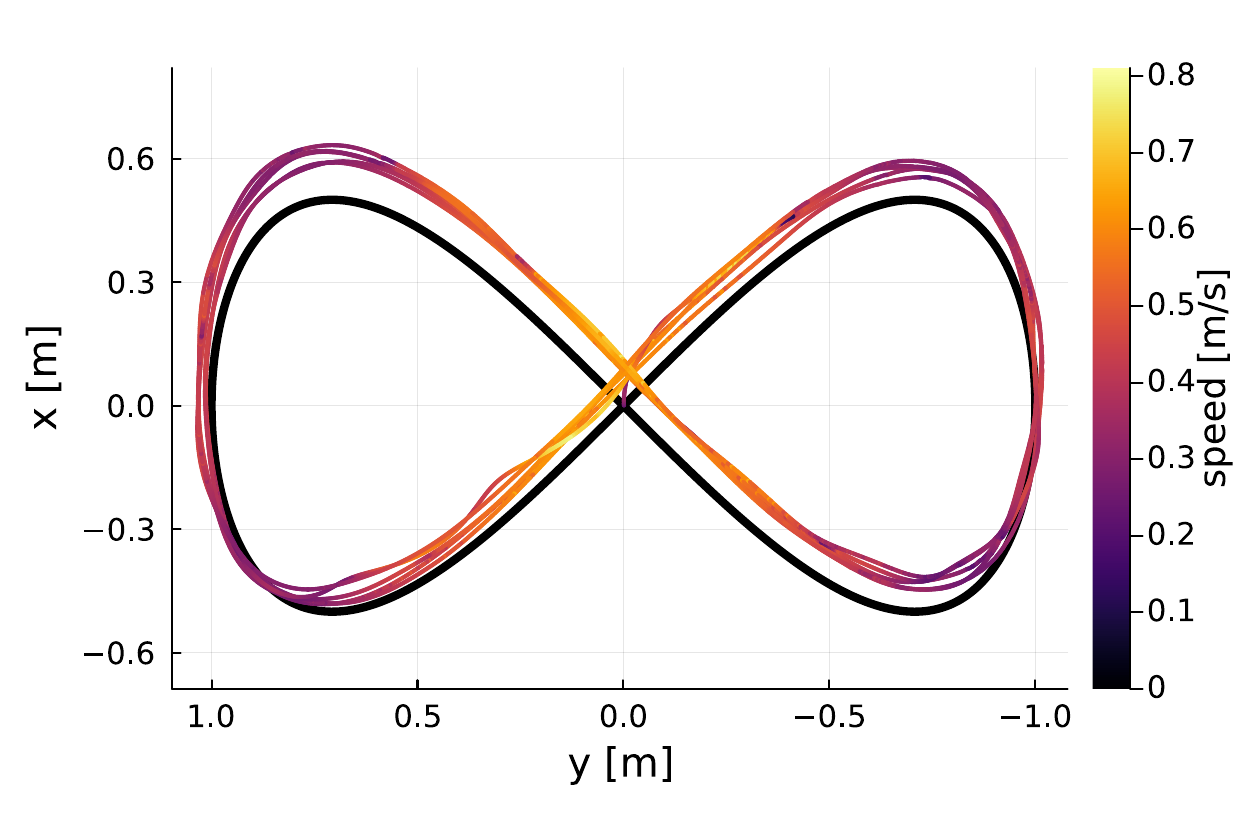}
        \caption{Slow ($T = \SI{15}{\second}$)}
        \label{fig:trajectory_tracking_lissajous_15}
    \end{subfigure}%
    \hfill
    \begin{subfigure}{0.33\textwidth}
        \centering
        \includegraphics[width=\linewidth]{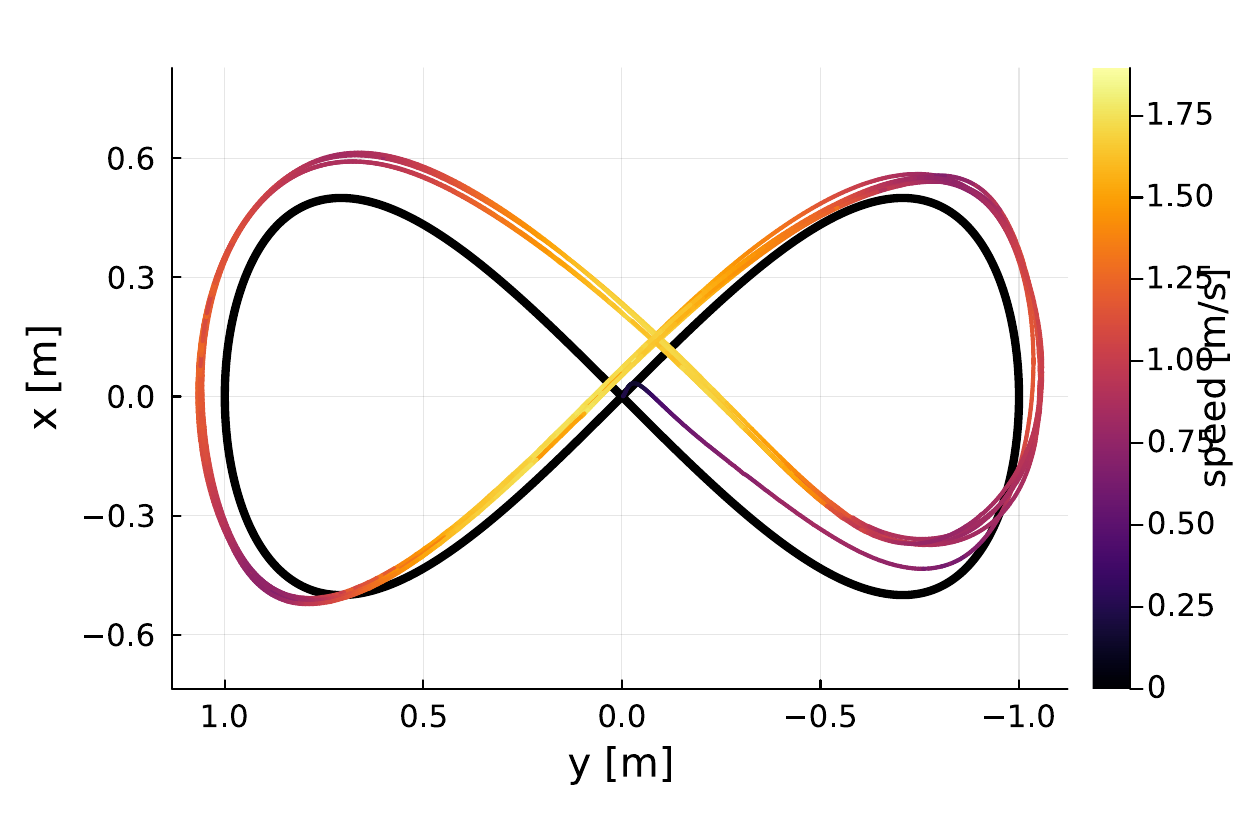}
        \caption{Normal ($T = \SI{5.5}{\second}$)}
        \label{fig:trajectory_tracking_lissajous_5.5}
    \end{subfigure}
    \hfill
    \begin{subfigure}{0.33\textwidth}
        \centering
        \includegraphics[width=\linewidth]{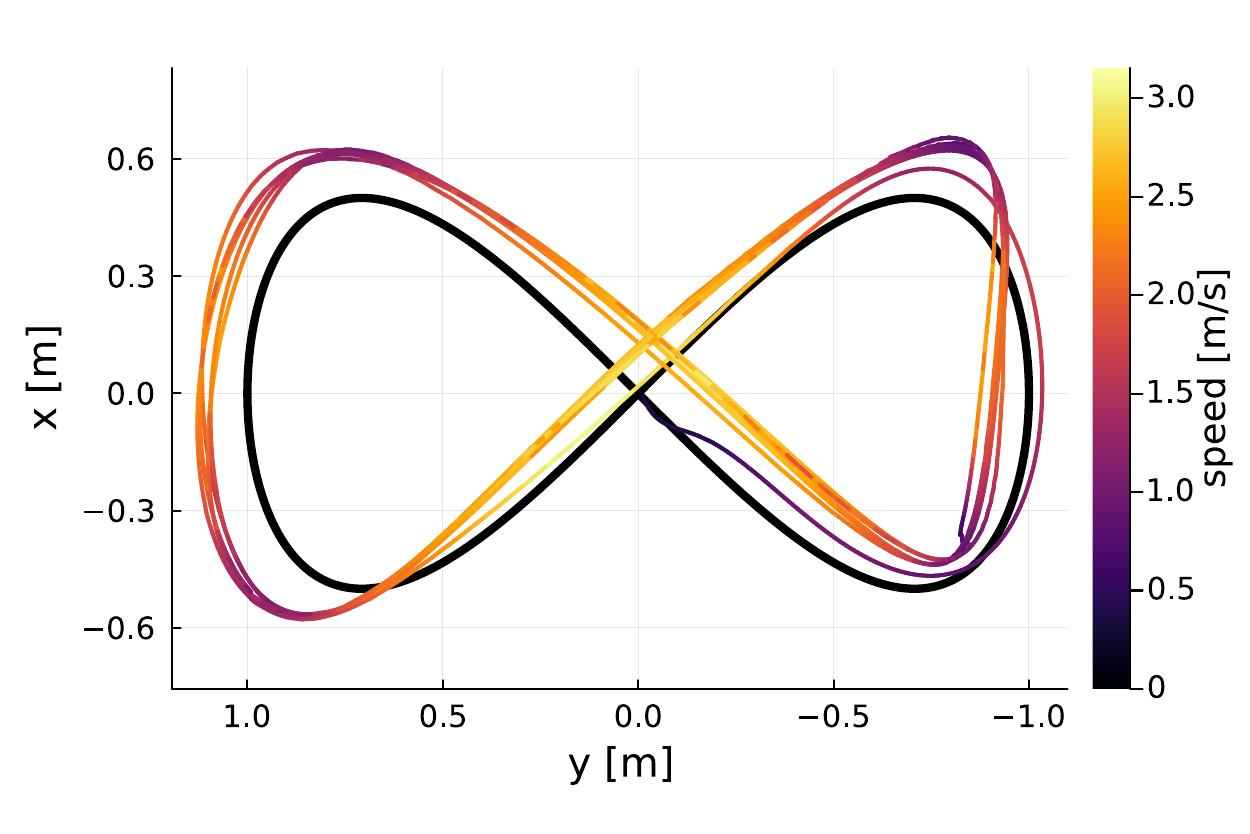}
        \caption{Fast ($T = \SI{3.5}{\second}$)}
        \label{fig:trajectory_tracking_lissajous_3.5}
    \end{subfigure}
    \caption{Real-world tracking of a Lissajous trajectory with different cycle times (reference trajectory in black).}
    \label{fig:trajectory_tracking_lissajous}
    \vspace{-10pt}
\end{figure*}

%% file: tables/trajectory_tracking_results.tex
\begin{table}[htb]
\def\arraystretch{1.3}
\centering
\begin{tabular}{c | c c | c c | c c}
Interval & \multicolumn{2}{c|}{Slow (\SI{15}{\second})} & \multicolumn{2}{c|}{Normal (\SI{5.5}{\second})} & \multicolumn{2}{c}{Fast (\SI{3.5}{\second})} \\
\hline
Controller & $\overline{e} $  & $\overline{e}_{xy} $ & $\overline{e} $  & $\overline{e}_{xy} $ & $\overline{e} $  & $\overline{e}_{xy} $  \\
\hline
\grayrow
PID           & $0.23$          & $0.22$          & $0.72$          & $0.72$          & $0.88$          & $0.87$ \\
Geometric~\cite{mellinger_minimum_2011}     & $\mathbf{0.06}$ & $\mathbf{0.04}$ & $\mathbf{0.16}$ & $0.16$          & $0.36$          & $0.36$ \\
\grayrow
Nonlinear~\cite{BrescianiniNonlinearController2013}     & $0.29$          & $0.11$          & $0.38$          & $0.32$          & $\infty$        & $\infty$ \\
INDI~\cite{smeur_indi_2016}           & $0.21$          & $0.21$          & $1.13$          & $1.13$          & $1.04$          & $1.04$ \\
\grayrow
Baseline~\cite{molchanov_sim--multi-real_2019} & $0.15$  & $0.13$ & $0.25$ & $0.24$ & $0.52$ & $0.50$ \\
Baseline~\cite{gronauer2022usingsimulation} & $\mathbf{0.06}$  & $0.05$ & $0.23$ & $0.21$ & $\infty$ & $\infty$ \\
\grayrow
\textbf{Ours} & $0.08$          & $0.08$          & $0.17$          & $\mathbf{0.15}$ & $\mathbf{0.24}$          & $\mathbf{0.22}$ \\
\hline
\end{tabular}
\caption{Real-world trajectory tracking error $\overline{e}$ (RMSE including $z$ in meter) and $\overline{e}_{xy}$ (RMSE excluding $z$ in meter) when tracking the Lissajous trajectory in Fig. \ref{fig:trajectory_tracking_lissajous} using different controllers.}
\label{table:controller_comparison}
\end{table}

%% file: sections/6_conclusion.tex
\section{Conclusion}
In this work, we presented an unprecedentedly fast end-to-end \gls*{rl} architecture for quadrotor control that directly outputs \glspl*{rpm} and can be trained to fly a real quadrotor in \SI{18}{\second} on consumer-grade laptops. The approach directly transfers to real-world platforms even without domain randomization. Compared to prior work, our approach leads to very reliable training behavior and does not require the cherry-picking of trained policies. Furthermore, we conducted a large ablation study validating our design decisions and found that our method is competitive with other classic and learned controllers. 
We open-source our approach and simulator setup to the community to democratize learning-based quadrotor control. Due to the curse of dimensionality, the design space (in terms of hyperparameters and other design decisions) of \gls*{rl}-based end-to-end quadrotor control is still sparsely explored and we believe that our experimental results constitute a foundation that future research can build upon. Our proposed training paradigm and the resulting highly optimized implementation allow for greatly reduced training times and hence more rapid iteration. 
Furthermore, the Crazyflie quadrotor uses open-source firmware, is widely available and inexpensive.

Future works will push the training speed and robustness as well as the tracking performance of learned low-level controllers through automatic hyperparameter optimization. Furthermore, we are interested in extending the policy to be adaptive to changing system or environment parameters like battery levels or wind, possibly using integral compensation \cite{xu2019integralcompensation} or meta-\gls*{rl} \cite{eschmann2021partially}.